\documentclass[a4paper, conference]{IEEEtran}
\IEEEoverridecommandlockouts
\usepackage{cite}
\usepackage{amsmath,amssymb,amsfonts}
\usepackage{graphicx}
\usepackage{textcomp}
\usepackage{xcolor}
\usepackage{multirow}
\usepackage{arydshln}
\usepackage{todonotes}
\usepackage{etoolbox}
\usepackage{setspace}
\usepackage[bottom]{footmisc}
\usepackage{placeins}
\usepackage{wrapfig}
\usepackage[export]{adjustbox}
\usepackage{amsmath}
\usepackage{algpseudocode}
\usepackage[ruled,vlined]{algorithm2e}
\usepackage{tikz}
\usepackage{relsize}
\usepackage{mathtools, nccmath}
\def\BibTeX{{\rm B\kern-.05em{\sc i\kern-.025em b}\kern-.08em
    T\kern-.1667em\lower.7ex\hbox{E}\kern-.125emX}}
    
\usepackage{url}

\patchcmd{\section}{\centering}{}{}{}

\begin{document}
    
\title{
   Knowledge Fusion Transformers for Video Action Classification
    \vspace{-0.3cm}
}

 \author{\IEEEauthorblockN
 {  Ganesh Samarth.$^{1,3}$,
    Sheetal Ojha$^{2,3}$,
    Nikhil Pareek$^{3}$
 }
 \\
 \IEEEauthorblockA{ Indian Institute of Technology Dharwad$^{1}$ ,
 Indian Institute of Techonology Roorkee$^2$,  UltraInstinct AI, India$^3$  \\
 170020030@iitdh.ac.in $^{1}$, sojha@mt.iitr.ac.in $^{2}$, nikhil@ultrainstinct.ai $^{3}$ \\
}
 }

\maketitle
\begin{abstract}
We introduce Knowledge Fusion Transformers for video action classification. We present a self-attention based feature enhancer to fuse action knowledge in 3D inception based spatiotemporal context of the video clip intended to be classified. We show, how using only one stream networks and with little or, no pretraining can pave the way for a performance close to the current state-of-the-art. Additionally, we present how different self-attention architectures used at different levels of the network can be blended-in to enhance feature representation. Our architecture is trained and evaluated on UCF-101\cite{soomro2012ucf101}, HMDB-51\cite{Kuehne11} and Charades \cite{sigurdsson2016hollywood} dataset, where it is competitive with the state of the art. It also exceeds by a large gap from single stream networks with no to less pretraining.

\end{abstract}


\section{Introduction} \label{sec:intro}

With the rapid growth and ubiquitous usage of CCTV cameras and other surveillance equipment, manual supervision has become less reliable and increasingly more difficult and hence paving the way towards automated solutions. Such a system requires real-time understanding and interpretation capabilities to classify actions with high reliability and efficiency. Video classification algorithms have also become critical for classifying and censoring video content on the internet for coherent content filtering.  

However, the most prevalent problem while analyzing videos is that surveillance clips are usually very long and hence it becomes necessary to capture both short-term and long term relationships simultaneously. A video may look entirely normal with few frames of criminal activity embodied in it. It becomes essentially important to recognize such activities to make surveillance automated.  

As humans, our brain has evolved to selectively concentrate on discrete aspects of the information provided by various receptors. Since birth, attention orients our reflexes to help us determine which events in our environment need to be attended to support our chances of survival. This concept of attention can be extended to the computer vision domain in the form of self-attention inherited in the network.

We understand the concept of time through a change in objects in space. As the theory of special relativity suggests that space and time cannot be separated, they are intertwined together to fabricate reality \cite{relativity2013}. Understanding the semantics of video content implies dealing with space as in the spatial dimension and time as in the temporal dimension, which are fabricated together to make the context of the video. It seems self-evident to use 3D convolutions \cite{6165309} which treat the volumetric content of the video as spatial local correlations and temporal non-local context simultaneously. 
More than one movement or step forges an action. The action involves a purpose, a target, and an agent that performs it. But unassociated movements like that of the camera or the objects in the background further intricate the task of video classification.

In this work, we present the Knowledge Fusion Transformers (KFT). 3D Inception blocks are used to extract spatio-temporal features, with attention applied at later levels of the network with lateral connections among the different layers of attention. This single stream network without pretraining on large video corpus of data is powerful enough to complete with existing state-of-the-art algorithms, making it computationally efficient in learning and usage. The pivotal idea is to treat the spatial and temporal dimensions in an unbiased way by applying attention across spatial-temporal features generated by 3d Convolution layers. We also present the ways in which the KFT can be integrated with other models to enhance their predictive power.

We experiment with different types of attention mechanisms on the HMDB-51 \cite{Kuehne11}, UCF-101 \cite{soomro2012ucf101} and Charades \cite{sigurdsson2016hollywood} datasets. KFT sets a new state-of-the-art in single stream networks with significant gains to previous systems in the literature.

The next section describes prior work in the domain as well as other related research. Section III presents an overview of our recommended approach. Section IV depicts our experimental setup and reports our results along with previous benchmarks with ablation experiments. The final section concludes the paper with the inference of our results and directions for future work.

\section{Related Works} \label{sec:related_works}

In the era of deep learning, video classification tasks has seen majorly two prominent approaches to solve the problem, namely single stream networks and two stream networks. The main difference lies in the treatment of temporal and spatial features. 
\begin{figure*}[!ht]
\centering
\includegraphics[width = 12cm]{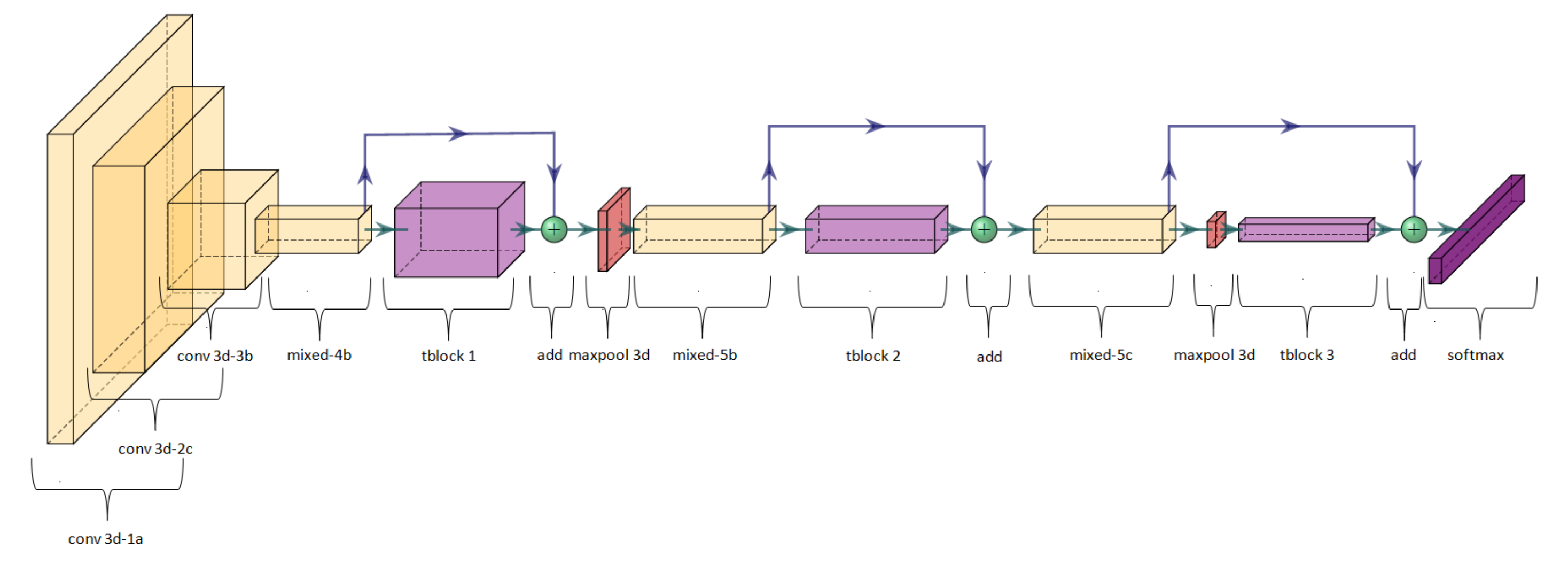}
\vspace{-0.1cm}
\caption{Knowledge Fusion Transformer General Architecture}
\label{fig:semigeneral}
\end{figure*}
Karpathy et al \cite{6909619} explored multiple ways to fuse temporal information from consecutive frames using 2D pre-trained convolutions. The idea was further extended by Du Tran et al \cite{tran2015learning} where authors used 3D convolutions on video volume to learn spatio-temporal features on a large scale supervised dataset and used SVMs for classification. \cite{donahue2014longterm} \cite{yao2015describing} explored the idea of using 2D CNNs on RGB or optical flow images as encoder blocks to capture spatial features and LSTMs as encoder blocks for temporal features, using end-to-end training for the entire architecture. The high performing image recognition networks were reused to extract features from each frame independently followed by an LSTM to capture temporal ordering and long-range dependencies.A major problem with this kind of an approach is that, in a single video clip, a Conv-LSTM model does not have the ability to perform such selective attention and treats all frames equally. Also, all sequential networks including LSTMs put extreme challenges on training large scale datasets.

For single steam networks, modelling long range temporal features was a major shortcoming. To overcome the above mentioned short-coming, videos were decoupled into different streams to capture spatial and temporal features separately. Simmoyan and Zisserman \cite{2streamZisserman} proposed that this decomposition is quite natural, where the spatial part, in the form of individual frames, carries information about scenes and objects depicted in the video, and the temporal part in the form of motion across the frames, which is essentially the motion of the observer and the objects. \cite{feichtenhofer2016convolutional} explored the ways in which the information from spatial and temporal features can be fused by a convolution layer rather than a softmax layer resulting in substantial saving with less parameters and pooling of abstract convolution features further boost the performance. 

 \cite{carreira2017quo} further refined the idea of using two steams networks for video recognition models by using Inception V1 network on RGB and optical flow images. Output from both of these steams were averaged out before the final prediction. The network also benefited from ImageNet \cite{imagenet_cvpr09} and Kinetics datasets \cite{kay2017kinetics} pre-training. The RGB frames were taken as input to the spatial stream and optical flow vectors were used for the temporal stream. Optical flow vectors essentially capture the motion of objects in between consecutive frames which can be caused by either the movement of the object or the observer. It uses pixel intensities as a function of space and time to trail motion trajectories. A popular method to generate optical flow vectors is the TV-L1 algorithm \cite{perez2013tv}, which uses the L-1 norm as the minimizing function. Pre-computed flow images were an essential part of the network, computationally inefficient for real-time applications. 
 
 The concept of fusion was taken further in \cite{feichtenhofer2019slowfast} where the authors used RGB frames with different frame rates as input to the temporal stream and spatial streams. The work also extended the idea of lateral connections from \cite{feichtenhofer2016convolutional} to combine the knowledge between the two streams effectively.

Attention was first introduced in \cite{bahdanau2014neural} solving the long-range dependency problem of RNNs and LSTMs in translating long sentences by giving relative importance to each word in the context vector. Thereafter it has also been used to support the non-sequential reasoning processes \cite{yi2019clevrer}. Local Self Attention \cite{parmar2018image} was a step towards improving the scalability of self-attention mechanisms for super-resolution models by limiting the memory position to a local neighborhood around the query and changing it with each of the queries position.

Attention incorporated in this work for video recognition is analogous to the one used for machine translation as a part of the transformer architecture \cite{vaswani2017attention} which suggested using only the attention mechanism for sequential tasks and completely eliminating the use of recurrent units and enhancing parallelization. 

One of the diverse uses of attention was by incorporating it with region proposal networks before attentions and using them on top of other networks like Inception V1 \cite{girdhar2019video}.

\cite{8578911} also introduce a similar idea to attentions by using generic non-local operations \cite{1467423} using (1X1x1) convolution operations by computing the response at a position as a weighted sum of the features at all positions to capture long term dependencies. Our work differs in the fundamental way as we believe video features should be represented as spatio-temporal features.

   \section{Proposed Approach}  \label{sec:proposal}

We build up the idea of using one steam networks from I3D \cite{carreira2017quo}, video action transformer \cite{girdhar2019video} and non local neural networks \cite{8578911}. These networks has following main drawbacks (i) Computational limitations of using optical flow (ii) Information from all the frames were not leveraged during self attention, hence non-local information was not utilized (iii) Diminishing gradient problems in larger networks were not addressed if fused with networks other than resnets \cite{7780459} (iv) Representing spatial and temporal features independently. 

\subsection{\textbf{KFT Networks }}
The KFT network can be described as a single stream architecture with main Inception 3D blocks with knowledge fusion transformers blocks in latter part of the network. Each KFT block also has lateral connections to other blocks. It is designed to classify a short clip into appropriate activity / act / class it belongs to. 

\subsubsection{Inception Modules}
The network is composed of 3D Inception modules resembling Inception V1 Networks ( \cite{7298594}, \cite{carreira2017quo} ), to encode the video into spatio-temporal features. It takes n frames (n=64,128) as input, with a stride of 2 or every alternate frame, with each frame been resized as 224*224, is processed by the model. The nomenclature and definition of inception blocks have resemblance with I3D blocks. 

The 3D inception trunk constitutes seven 3D Inception blocks along with 3D Conv and 3D MaxPool layers. The final features are extracted using an Average Pooling layer. Spatial compression is performed using a set of two 3d Convolution layers using kernel sizes of {$1 \times 4 \times 4$} and {$1 \times 5 \times 5$}.. input of T*H*W with number of channels = 3 is changed into spatio-temporal feature of T/4 * H/16 * W/16 with number channels = 832. After T-block1 and T-block2, we again have inception blocks namely mixed-5b and mixed-5c. 

 \begin{table}[!ht]
    \centering
    \resizebox{7cm}{!}{
    \begin{tabular}{|c|c|c|}
    \hline
        Stage & Layer-params & Output sizes \\ \hline
        Raw Clip &  & $64 \times 3 \times 224^2$ \\ \hline
        Data Layer & stride 2, $1^2$ & $64 \times 3 \times 224^2$ \\ \hline

        \multirow{2}{*}{Conv3d-1a} & {$7 \times 7^2$} & \\  & stride 2, padding 3 &{$64 \times 16 \times 112^2$} \\ \hline
        
        \multirow{2}{*}{MaxPool3d-2a} & {$1 \times 7^2$} & \\ & stride 1,2,2 & {$64 \times 16 \times 56^2$} \\ \hline
        
        \multirow{2}{*}{Conv3d-2b} & {$1 \times 1^2$} & \\ & stride 1 & {$64 \times 16 \times 56^2$} \\ \hline
        
       \multirow{2}{*}{Conv3d-2c} & {$3 \times 3^2$} & \\ & stride 1 & {$192 \times 16 \times 56^2$} \\ \hline
       
       \multirow{2}{*}{MaxPool3d-3a} & {$1 \times 3^2$} & \\ & stride 1,2,2 & {$192 \times 16 \times 28^2$} \\ \hline
       
        \multirow{3}{*}{Mixed-3b \texttimes \hspace{0.1cm} \textbf{2} }& {$1 \times 3^2$} & \\ & {$1 \times 1^2$} &\\ & MaxPool3d - {$3 \times 3^2$} & {$256 \times 16 \times 28^2$} \\ \hline
    
        \multirow{2}{*}{MaxPool3d-4a} & {$3 \times 3^2$} & \\ & stride 1,2,2 & {$480 \times 8 \times 14^2$} \\ \hline
        
        \multirow{3}{*}{Mixed-4b \texttimes \hspace{0.1cm} \textbf{5}} & {$1 \times 3^2$} & \\ & {$1 \times 1^2$} &\\ & MaxPool3d - {$3 \times 3^2$} & {$832 \times 8 \times 14^2$} \\ \hline
        
        \multirow{2}{*}{AvgPool3d} & {$1 \times 7^2$} & \\ & stride 1,2,2 & {$832 \times 8 \times 8^2$} \\ \hline
        \multirow{2}{*}{KFT-block1} & heads 2, num. attentions 2 & \\ & linear-params 832 & {$832 \times 8 \times 8^2$}\\ \hline
        \multirow{2}{*}{MaxPool3d-5a} & {$2 \times 2^2$} & \\ & stride 2 & {$832 \times 4 \times 4^2$} \\ \hline
        \multirow{3}{*}{Mixed-5b }& {$1 \times 3^2$} & \\ & {$1 \times 1^2$} &\\ & MaxPool3d - {$3 \times 3^2$} & {$1024 \times 4 \times 4^2$} \\ \hline
        \multirow{2}{*}{KFT-block2} & heads 4, num. attentions 4 & \\ & linear-params 1024 & {$1024 \times 4 \times 4^2$}\\ \hline
        \multirow{3}{*}{Mixed-5c }& {$1 \times 3^2$} & \\ & {$1 \times 1^2$} &\\ & MaxPool3d - {$3 \times 3^2$} & {$1024 \times 4 \times 4^2$} \\ \hline
        \multirow{2}{*}{MaxPool3d-6a} & {$2 \times 2^2$} & \\ & stride 2 & {$1024 \times 2 \times 2^2$} \\ \hline
        \multirow{2}{*}{KFT-block3} & heads 4, num. attentions 6 & \\ & linear-params 1024 & {$1024 \times 2 \times 2^2$}\\ \hline
        \multirow{2}{*}{Conv3d-logits} & {$2 \times 2^2$} & \\ & stride 2 & {$101 \times 1$} \\ \hline

    \end{tabular}}
    \vspace{0.4cm}
    \caption{KFT architecture description}
    \label{tab:Inception_table}
\end{table}

\begin{figure*}[!ht]
\centering
\includegraphics[width = 8cm]{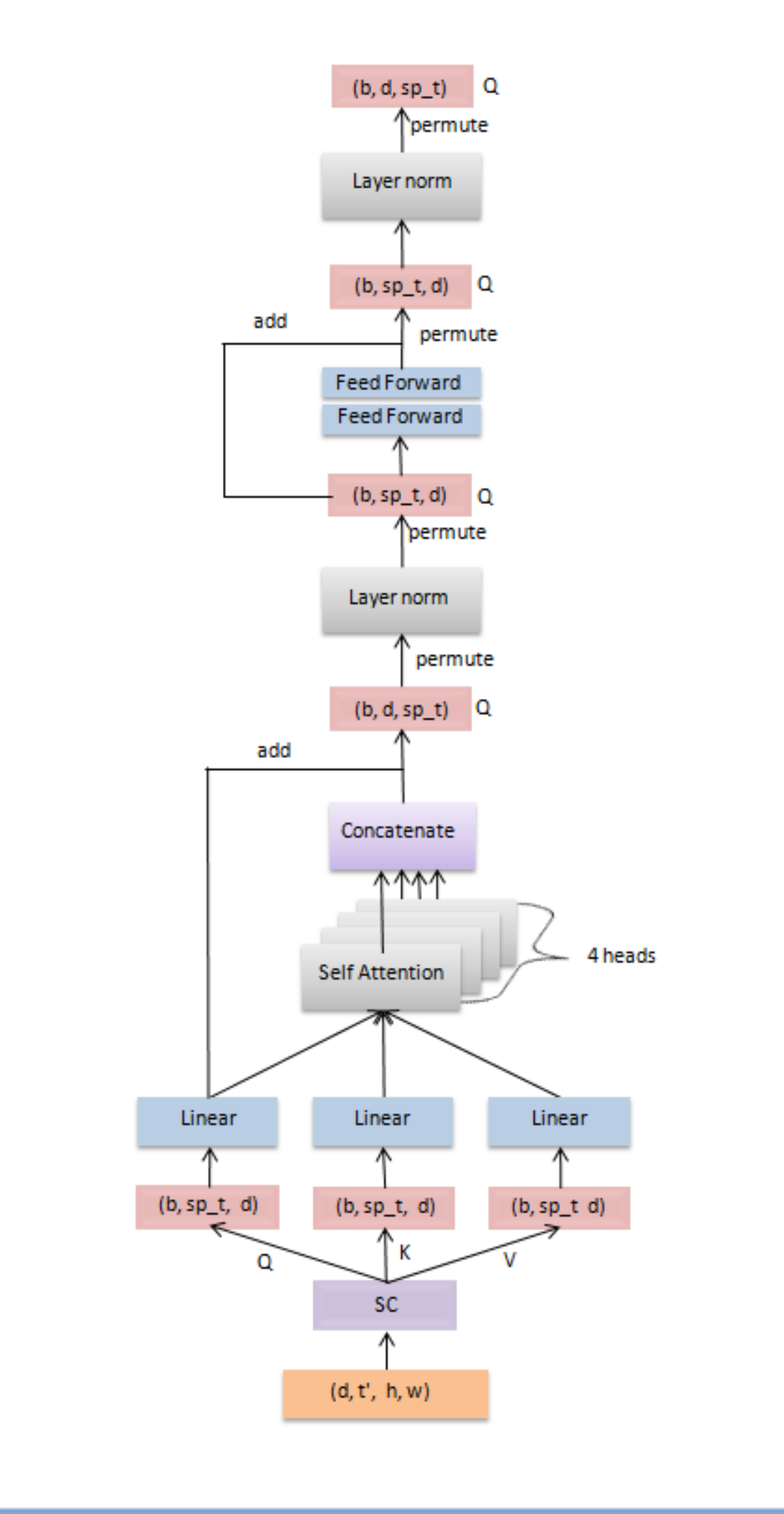}
\vspace{0.1cm}
\caption[Attention Mechanism:]{  In each KFT block of the network, the input size is (d,t,h,w), where d represent the feature vector in  time and space dimension. We apply spacial compression (SC) in case of 1D and 2D KFT block.They are modeled by 2 convolution blocks. We then reshape the space and time dimension into one space, sp-t = t*h*w.After taking linear projections as standard transformer, we get out query(Q), key (K), and value (v) pairs. sp-t is again divided to perform multihead attention, followed by concatenation of layers}

\label{fig:attentionmech}
\end{figure*}

\subsubsection{Attention mechanisms}
Our self-attention mechanism is inspired from Video Action Transformer Model \cite{girdhar2019video} and Transformer Model \cite{NIPS2017_7181}. We have explored different Attention Mechanisms on extracted features, and affect of number of heads and self-attention layers vs the placement of transformer, and how lateral connection boosts the training and inference. 
Self attentions in neural networks have proven to be very beneficial technique applied across domain tasks (\cite{NIPS2017_7181}, \cite{xu2015attend}, \cite{carion2020endtoend} ). It started with applications in sequence to sequence tasks, but has gained a lot of applications in computer vision in recent time. The main idea of the original seq2seq model is to enhance feature weights by comparing features with other features in the sequence. The idea is similar to using non-local blocks \cite{8578911} with difference been the computation of relative weights among features. 

To generate  query, key and value (Q ,K ,V), features from inception modules were either preprocessed or projected linearly. "Scaler Dot Product" attention (multiplicative) with multiple heads is incorporated to generate a richer representation of the features. Output is also positional encoded to incorporate positional information. \\

\begin{equation}
    Attention(Q,K,V) = softmax(\frac{QK^T}{\sqrt{d_k}}) 
   \end{equation}
    
We experiment with three different types of attention mechanisms retaining the same basic operation by modifying the query, key and value for each type of attention.\\

\textbf{Attention-1D} This reformed attention mechanism is inspired by the work outlined in \cite{girdhar2019video}. The shape of the obtained features are represented as {$C \times D \times H \times W$}, where C denotes the number of channels, D denoted the depth (temporal dimension) , H and W represent the height and width of the feature space (spatial dimensions). We improve upon the previous work to reduce the spatial layout by using Spatial Compression layers in the form of 3D Convolutions resulting in a feature vector of shape {$C \times D \times 1 \times 1$}. Linear projections of this feature vector directly taken as the key and value components. The middle frame from this vector is extracted and taken as the query. The intuition behind this approach is when attention is applied upon the middle frame, a context is generated establishing importance of each object in every frame.
The extracted vector of shape {$C \times 1 \times 1 \times 1$} is expanded to the original size of {$C \times D \times 1 \times 1$}. The mechanism is stacked with multiple attention heads and the concatenated features are taken as the updated query as described in the original transformer model \cite{vaswani2017attention}. The new query is again used to generate a deeper contextual understanding by further applying attention. This unit is described in Figure \textbf{\ref{fig:attentionmech}} \\

\textbf{Attention-2D} Building on the previous model, instead of using just the middle frame, the query, key and value are taken as the entire feature sequence after applying spatial compression. By using the entire sequence, relative context of  each feature with respect to the other is generated rather than with respect to one frame and hence a richer representation. The updated query is refined through several layers of attention. If the shape after spatial compression is taken to be {$C \times D \times 1 \times 1$} the output is flattened to be of shape {$C*D \times 1 \times 1 $} and passed through two linear layers followed by softmax to generate predictions .\\

\textbf{Attention-3D}  The Inception trunk is comprised of 3D Convolution layers, hence all features generated encode information in a spatio-temporal sense. Therefore, in this attention mechanism, the spatial compression layers are avoided.
Given a feature representation of shape {$C \times D \times H \times W$} the vector is directly flattened to {$C \times D*H*W$}.  The modified attention procedure is to capture spatio-temporal inter-dependencies which would otherwise be lost while performing spatial compression. The query, key and value are taken to be the linear projections of this flattened vector. Since attention is shape invariant, and all the feature are retained, the classical I3D tail is incorporated. To efficiently capture a combination of both local and non-local dependencies among spatio-temporal features, the inception and transformer blocks are interspersed with each other as in Fig. \ref{fig:semigeneral}.  \\

\begin{table*}[!ht]
    \centering
    \begin{tabular}{|c|c|c|c|}
     \hline
     
    \multirow{2}{2cm}{Architecture}     & \multirow{2}{1.5cm}{UCF-101} & \multirow{2}{1.75cm}{HMDB-51} & \multirow{2}{1.5cm}{Charades (mAP)} \\ & & & \\
    
    \hline 
       
        C3D \cite{tran2015learning} & 51.6 & 24.3 & -\\ \hline
        3D-SqueezeNet \cite{kopuklu2019resource} & 74.9 & 55.6 & -\\ \hline
        LSTM \cite{6909619} & 81 & 36 & -\\ \hline
        3D-Fused \cite{feichtenhofer2016convolutional} & 83.2 & 49.2 & - \\ \hline
        Two-Stream \cite{simonyan2014two} & 83.6 & 43.2 & 18.6 \\ \hline
        I3D (RGB) \cite{carreira2017quo}& 84.5 & 49.8 & 32.9 \\ \hline
        iDT+ HSV \cite{peng2016bag} & 87.9 & - & -\\ \hline
        I3D (Flow) \cite{carreira2017quo} & 90.6 & 61.9 & 35.2 \\ \hline
        LTC \cite{varol2017long} & 91.7 & 64.8 & -\\ \hline
        SlowFast \cite{feichtenhofer2019slowfast} & 92.1 & -& 42.1 \\ \hline
        Two-Stream (RGB+Flow) I3D \cite{carreira2017quo} & 93.4 & 66.4& -\\ \hline \hline
        KFT-1D & 87.4 & 55.3 &\\ \hline
        KFT-2D & 88.4 & 56.5 &\\ \hline
        KFT-3D & \textbf{92.4} & \textbf{67.2} & \textbf{42.3}\\ \hline
        
    \end{tabular}
    \vspace{0.2cm}
    \caption{Knowledge Fusion Transformer Performance Metrics}
    \label{tab:semi_metrics}
    \end{table*}

\subsubsection{Residual Connections}
Since the network proposed in very deep with 11 to 18 layers of self attention blocks (2-4-6 or 6 for each transformer block), residual connections are introduced to avoid the vanishing gradient problem. Also, we wanted to extract new features in spatio-temporal dimension, but wanted to keep the refined features after KFT blocks. Two types of residuals are experimented with - concat and additive.
Concatenating the input after each transformer block didn't give a considerable boost to the performance since the features from KFT blocks were tempered in a wrong way. But with additive connections, we got a performance boost of ~2\% in accuracy on UCF-101.

\section{Experiments} \label{sec:exp}

\textbf{Datasets :} UCF-101 \cite{soomro2012ucf101} which totally consists of about 13k videos, with 3.7k videos for testing and 9.8k videos for training spread across 101 classes. HMDB-51 \cite{Kuehne11} dataset, with about 5k training and 2k validation clips. We report the top1 and top5 mean accuracies obtained by training on all the three splits. Charades has 9.8k training videos and 1.8k validation videos distributing among 157 classes in a multi-label classification setting of longer activities spanning 30 seconds on average. Performance is measured in mean Average Precision (mAP).

\textbf{Training:} All networks were trained on 4X GTX-1080 Ti  GPus by using ImageNet pre-training only on the starting 3D Inception blocks. For the remaining layers of our model (eg, Transformers, inception blocks, etc) from scratch and then trained the model end-to-end for both of the datasets. We also used a combination of batch norm and layer norm for normalizing 3D convnets and transformer blocks respectively. We used synchronized SGD training from \cite{goyal2017accurate} with an effective batch size of 32, or 8 batch size per GPU, and momentum of 0.9 were used. We also used SyncBatchNorm implementation in pytorch since most of our implementation were from scratch. Most of the models reached their peak accuracy <35 episodes of training. We used a starting learning rate of 0.01 with cosine learning rated for UCF-101 and HMDB-51 and step LR decreasing policy where the LR was decreased by a factor of 0.1 if the validation error saturated. We used cross entropy loss functions for training UCF-101 and HMDB, and used Charades training strategy from \cite{feichtenhofer2019slowfast} by using per class Sigmoid output to account for the multi class nature with binary entropy loss function.

\textbf{Data Augmentation :} 128 frames are sampled from each video randomly and 32 frames at equal intervals are selected. The spatial random cropping of  [224,224] with scale jittering range of [255,320] and random flipping. As mentioned in \cite{girdhar2019video}, the importance of data augmentation in videos, helped in removing over-fitting with significant performance improvements. All videos undergo mean normalization and are trained on 3 different splits.

\subsection{Main Results}

In this section we will compare the performance of few state-of-the-art techniques with our 3 implementation stated in previous section. We  will also, compare the effect of using different number of attention heads and self-attention blocks.

\subsubsection{Comparison with the State-of-the-Art} 

We show a comparison of the performance of our model and previous state-of-the-art models in table, on UCF-101, HMDB-51 and Charades. We include the results of models with or without pretraining on ImageNet.

We will be comparing our results with methods that have not used any additional data (eg, Kinetics pretraining), other than ImageNet pretraining for comparing our results. We were not able to pretrain on Kinetics because of resource limitation within the organization, but we surely believe that pretraining on Kinetics will boost our performance. 

Two of the best performing and landmark models on these datasets were I3D and SlowFast Models. We trained both of the networks with similar configurations as ours. The benchmarks were generated using mean accuracy over the standard train and test splits. Our model outperformed both of the models. The model also converged faster, with reaching the top accuracy in lesser episodes.

\subsection{Ablation Experiments}

Several experiments were conducted to determine the best settings for the proposed model which case be broken down into these points (i) Different input frame lengths with temporal strides (ii) Number of transformer heads and blocks (iii) Different settings for residual/lateral connections (iv) Batch norm and layer norm settings at different locations of Architecture (v) and Model Initialization for training from scratch.

Spatial Dimensions were varied between 224 to 256 and temporal dimensions from 8 to 64 frames. The results are indifferent to the spatial dimensions however there is a gradual increase in performance from 8 frames to 32 after which the model saturates. Very little performance gains can be seen from 32 frames to 64 frames. From our experiments, temporal strides depends on the nature of dataset and activity. We experimented with 2,4,8 and found out using temporal stride of 2 gives optimum results. 

Number of heads and blocks in each attention block is also varied from 2 to 8 along with the number of  attention block in each transformer block. Our experiments are tabulated in [table]. We found out increasing number of transformer blocks as we go deeper into the network has optimum performance. The performance improves very slightly if we increase number of blocks in KFT blocks with larger impact on performance.

We tried 2 settings for lateral connections among attention layers. (i) Adding connections among inception blocks (ii) Adding residuals from previous KFT block to next KFT block. The former didn't perform well, but the later further gave performance gains, both in terms of accuracy (+) and in terms of training time(-).

To avoid over-fitting as well as vanishing and exploding gradient problems, different combinations of Batch Normalization and Layer Normalization across different layers are experimented with. The best performance is achieved while using BatchNorms after Conv-layers and LayerNorms after Attention. Kaiming \cite{he2015delving} and constant weight initialization was experimented with for the feed-forward layers in the transformer network.

    \section{Conclusion and Further Research}

The relationship between space and time will always be complex. We tried our best it to unravel it a bit. As we explained that time and space are inseparable dimensions, hence treating them similar to spatial dimensions makes more sense with added complexity of one more dimensions. Also, we believe that addition of KFT blocks, similar to a normal neural network block, to any other network component will further improve those models and pave way for new research of improved and hybrid transformer blocks in Video Recognition Architectures. It also trains faster and achieves state-of-the-art in the limited episodes of training. As it is evident that with pretraining on a larger video dataset improves results drastically, our model have potential to be improved further. 

As stated, next steps can be to train it on larger dataset or it can be make two steam network to further enhance its performance. It would be great if we could have added these comparisons but due to limited resources and time, we could not. Also, a better architecture exploration can be performed to combine it with more out-of-box ideas. Region proposal networks (RPN) \cite{Ren2015:FasterRCNN}  can also be added into the network to enable action localization and detection.
    
\small{
\bibliographystyle{IEEEtran}
\bibliography{main}
}
\end{document}